\documentclass{article}

\usepackage{iclr2027_conference,times}

\usepackage{amsmath,amsfonts,bm}

\def\eqref#1{equation~\ref{#1}}
\def\1{\bm{1}}

\DeclareMathAlphabet{\mathsfit}{\encodingdefault}{\sfdefault}{m}{sl}
\SetMathAlphabet{\mathsfit}{bold}{\encodingdefault}{\sfdefault}{bx}{n}

\usepackage{amsmath}
\usepackage{amssymb}
\usepackage{booktabs}
\usepackage{graphicx}
\usepackage{hyperref}
\hypersetup{hidelinks}
\usepackage{url}
\usepackage{fvextra}
\usepackage{placeins}
\DefineVerbatimEnvironment{PromptTemplate}{Verbatim}{%
    fontsize=\footnotesize,breaklines=true,breakanywhere=true,
    breaksymbolleft={},samepage=false,
    frame=single,framesep=2mm}

\newcommand{\meanstd}[2]{#1{\scriptsize$\pm$#2}}

\title{VH-GPO: Verified Handoff for\\
LLM Agent Reinforcement Learning}

\author{Haoze Li\textsuperscript{1}, Peng Zhou\textsuperscript{2,*},
Peiyuan Wu\textsuperscript{2}, Weijie Zhao\textsuperscript{2},
Yinan Xin\textsuperscript{2}}

\iclrfinalcopy

\begin{document}

\maketitle
% Override the conference publication header set by the template in \maketitle.
\lhead{Preprint}
\footnotetext[1]{Beihang University. Email: \texttt{iceblwdzs@gmail.com}.}
\footnotetext[2]{Luxi Tech, Shenzhen, China.}
\begingroup
\renewcommand{\thefootnote}{\fnsymbol{footnote}}
\footnotetext[1]{Corresponding author: Peng Zhou. Email: \texttt{zp@luxitech.cn}.}
\endgroup
\setcounter{footnote}{2}

\begin{abstract}
Reinforcement learning (RL) plays a key role in training large language models
(LLMs) to solve complex tasks through outcome-based feedback.
Group-relative RL methods such as GRPO train
language-model agents by comparing the outcomes of multiple rollouts for the
same task. This mechanism is effective when outcomes vary, but if every
trajectory in a group fails with the same reward, relative advantages collapse
toward zero and the group supplies no useful learning signal. When the
trainable agent's rollouts fail to produce any successful trajectory, one
possible solution is to introduce a frozen teacher policy. We refer to the
trainable agent as the student. The teacher attempts recovery from a state
reached by the student, and its continuation is retained only if the
environment verifies success. Yet allowing the teacher to complete
the remaining task makes the entire recovered suffix teacher-generated,
turning supervision into off-policy imitation and weakening a key benefit of
RL: learning on the state distribution induced by the student. We therefore
ask: after a teacher has repaired a student failure, when should control return
to the student? Our core insight is that the teacher should act only long
enough to cross what the student cannot yet handle, then hand back control so
that successful experience is again student-generated. We formalize this idea
as \emph{outcome-verified teacher--student handoff} and instantiate it in
Verified Handoff for Group Policy Optimization (VH-GPO). Given a verified
teacher recovery, VH-GPO returns control to the student at the earliest
independently confirmed handoff point found within a bounded search budget.
With Qwen2.5-1.5B,
VH-GPO increases success rates over GRPO from 59.38\% to 94.79\% on ALFWorld
and from 71.62\% to 77.34\% on WebShop. Improvements also hold with the 7B
student, for which additional teacher and student output tokens amount to
1.77\% and 2.61\% of regular student rollout output tokens on ALFWorld and
WebShop, respectively. Ablation studies validate the design of
VH-GPO and demonstrate the effectiveness of its key components.
The teacher need not solve every problem for the student; instead, it should
provide just enough support when the student struggles to help it find a way
forward and gradually learn to solve problems independently.

\end{abstract}

\section{Introduction}
\label{sec:introduction}

Reinforcement learning (RL) has become an important component of large language
model (LLM) post-training, improving instruction following and reasoning
\citep{ouyang2022instructgpt,deepseekai2025r1}. Beyond text generation, LLM
agents reason over observations and issue actions to external environments,
including web browsing~\citep{nakano2022webgpt,yao2023react}. RL is
particularly attractive for improving such agents because the
agent learns from trajectories sampled by its own evolving policy. The
resulting supervision reflects the states, actions, and errors that the policy
will actually encounter, rather than only those chosen by a fixed demonstrator.
In long-horizon environments, however, outcome supervision is often extremely
sparse: an agent may execute many locally plausible actions and still receive no
terminal reward. This problem is especially acute for tasks that require
several dependent subgoals, such as manipulating multiple objects in ALFWorld
\citep{shridhar2021alfworld} or satisfying multiple product constraints in
WebShop~\citep{yao2022webshop}.

Group-relative policy optimization methods such as GRPO
\citep{shao2024deepseekmath} learn by comparing the outcomes of multiple
rollouts for the same task. When every rollout fails with the same terminal
reward, however, the group-relative advantages collapse to zero, leaving
no outcome-based policy-gradient signal to guide improvement on that task.
Such groups expose failures of the current policy without providing a
successful alternative to learn from. If these failures recur, the
corresponding capability gaps can persist even as the policy improves on
tasks with informative reward contrast. DAPO maintains informative training
batches by filtering out all-success and all-failure groups through dynamic
sampling~\citep{yu2025dapo}. When zero-advantage groups are frequent, as can
occur early in training on sparse-reward tasks, dynamic sampling requires
additional rollouts to replace discarded groups and fill each training batch,
increasing rollout cost. Yet filtering alone does not provide corrective
targets for the discarded failures. Fine-grained grouping and graph-based
credit assignment improve attribution within sampled experience
\citep{feng2025gigpo,cheng2026graphgpo}, yet cannot supply a successful
continuation when none is present. This motivates a mechanism that turns
selected zero-signal failures into verified learning opportunities.

When the current policy cannot produce a successful continuation, a natural
recovery mechanism is to invoke a frozen auxiliary policy from a failure state
that the policy has already reached. The continuation is retained only if the
environment verifies terminal success~\citep{xia2026ovcsd}, thereby turning an
otherwise uninformative failure into a grounded positive target. From this
point onward, we call the trainable policy the \emph{student} and the frozen
recovery policy the \emph{teacher}. A full teacher continuation can convert a
failed trajectory into a verified success. However, the longer the teacher
remains in control, the more student-generated RL experience is replaced by
off-policy imitation on the teacher's state--action distribution.

This raises the question of when a recovering teacher should return control
to the student. We formulate
\emph{outcome-verified teacher--student handoff} as a training principle:
the teacher corrects a student failure, then hands back control once
independent student continuations verify successful completion.

Based on this principle, we propose Verified Handoff for Group Policy
Optimization (VH-GPO). For an all-failure, zero-signal group, VH-GPO invokes a
frozen teacher from a student-reached state and retains its recovery only if
the environment verifies success. It then tests states along that recovery
to identify, within a bounded search budget, the earliest tested state from
which student success is independently confirmed. The resulting short teacher
bridge and verified student continuation provide three auxiliary training
signals: state-aligned action preferences, teacher-bridge supervision, and
student-resume supervision. These signals complement the unchanged root GRPO
update. The teacher is used only during training; the student acts
independently at inference time.

Experiments on ALFWorld and WebShop with Qwen2.5-Instruct students at 1.5B and
7B scales show that VH-GPO consistently improves the
corresponding GRPO baseline. With the 1.5B student, success increases from
$59.38\%$ to $94.79\%$ on ALFWorld and from $71.62\%$ to $77.34\%$ on WebShop.
With the 7B student, success increases from $78.12\%$ to $95.57\%$ on ALFWorld
and from $71.61\%$ to $79.69\%$ on WebShop. These gains require little additional
generation: in the 7B runs, all additional teacher and student generation
totals just $1.77\%$ of the tokens generated by regular student rollouts on
ALFWorld and $2.61\%$ on WebShop.

Our contributions are threefold:
\begin{itemize}
    \item We highlight the lack of corrective learning targets in all-failure,
    zero-signal groups and formulate \emph{outcome-verified teacher--student
    handoff} as a training principle: the teacher should provide only
    the support needed to restore the student's ability to complete the task,
    then return control once independent completion is verified.

    \item We introduce VH-GPO, which implements this principle through
    outcome-verified teacher recovery and bounded search for the earliest
    tested, independently confirmed student handoff. The resulting relays
    supply state-aligned action preferences, teacher-bridge supervision, and
    student-resume supervision alongside the unchanged root GRPO objective.

    \item Experiments on ALFWorld and WebShop show that VH-GPO improves overall
    task success over GRPO and competitive group-based RL baselines at both
    1.5B and 7B scales, with modest generation overhead: additional teacher
    and student generation totals only $1.77\%$ and $2.61\%$ of root rollout
    output tokens on the two benchmarks, respectively, in the 7B runs.
\end{itemize}

\section{Related Work}
\label{sec:related_work}

\paragraph{Reinforcement learning for language-model agents.}
RL trains language-model agents for web interaction, device control, and
search~\citep{qi2025webrl,bai2024digirl,jin2025searchr1}, supported by
environment and training frameworks such as AgentGym, AgentGym-RL, and
Agent Lightning~\citep{xi2024agentgym,xi2025agentgymrl,luo2025agentlightning}.
PPO and ArCHer provide actor--critic approaches to sequential
decision-making~\citep{schulman2017ppo,zhou2024archer}, while RAGEN examines
multi-turn training instabilities, including reward-variance
collapse~\citep{wang2025ragen}. Critic-free methods include REINFORCE
leave-one-out and GRPO~\citep{ahmadian2024back,shao2024deepseekmath};
REINFORCE++, Dr.\ GRPO, and GSPO refine normalization or importance
weighting~\citep{hu2025reinforcepp,liu2025drgrpo,zheng2025gspo}.
Group-relative estimation depends on within-task reward variation. DAPO
filters out all-success and all-failure groups through dynamic
sampling~\citep{yu2025dapo}.

\paragraph{Fine-grained credit assignment for multi-turn agents.}
Process supervision supplies intermediate feedback through human or
automatically generated step labels~\citep{lightman2023verify,wang2024mathshepherd}.
MCTS-based preference learning, ReST-MCTS*, and Agent Q use search to
construct finer-grained training
signals~\citep{xie2024mctsdpo,zhang2024restmcts,putta2024agentq}.
VinePPO estimates advantages using Monte Carlo continuations from
intermediate states~\citep{kazemnejad2025vineppo}; VH-GPO uses such
continuations to verify student resumability. For interactive agents, GiGPO
combines episode-level advantages with same-state action
groups~\citep{feng2025gigpo}, HGPO groups steps by shared historical
context~\citep{he2026hgpo}, and GraphGPO assigns credit through a
state-transition graph and distance to a verified goal~\citep{cheng2026graphgpo}.

\paragraph{Distillation for LLM agents.}
STaR, ReST, ReST$^{\mathrm{EM}}$, RAFT, and ReST meets ReAct learn from
successful or feedback-selected model
outputs~\citep{zelikman2022star,gulcehre2023rest,singh2024restem,dong2023raft,aksitov2023restreact},
while self-imitation learning reinforces past good decisions through an
off-policy actor--critic objective~\citep{oh2018sil}. Teacher supervision
can supply additional targets, including rationales in Distilling
Step-by-Step~\citep{hsieh2023distilling}. Generalized Knowledge Distillation
and MiniLLM reduce distribution mismatch using student-generated sequences
and teacher feedback~\citep{agarwal2024opd,gu2024minillm}, but matching
teacher predictions alone does not verify environment success.
Outcome-Verified Comparative Self-Distillation
(OVCSD)~\citep{xia2026ovcsd} organizes failed student rollouts into a prefix
tree, invokes a teacher from selected student states, and retains successful
continuations for divergence-level comparison and suffix distillation.
Full-suffix distillation can include segments that the student could already
complete independently. VH-GPO uses an independently confirmed student
handoff to determine where teacher supervision ends.

\paragraph{Selective intervention and recovery.}
DAgger and SafeDAgger study learner-state supervision and selective expert
queries~\citep{ross2011dagger,zhang2016safedagger}. HG-DAgger and ThriftyDAgger
regulate recovery interventions~\citep{kelly2019hgdagger,hoque2021thriftydagger}.
Go-Explore and Jump-Start RL exploit intermediate
states~\citep{ecoffet2021goexplore,uchendu2023jsrl}; Reflexion uses reflection
without parameter updates~\citep{shinn2023reflexion}.
ActGuide-RL~\citep{ji2026actguide} minimizes action-plan guidance after
all-failure rollouts and trains guided and unguided trajectories jointly.
DASH-OPD~\citep{xia2026dashopd} uses discrepancy-based hysteresis for
teacher--student switching in on-policy distillation. VH-GPO instead verifies
student handoff through independently sampled successful continuations
to learn from all-failure groups in group-relative RL.

\section{Method}
\label{sec:method}

\subsection{Problem Setup}
\label{sec:problem_setup}

We consider episodic agent tasks with environment-verifiable outcomes. In our
training setup, each episode is capped at $H$ environment interactions. After
$t$ interactions, the remaining interaction budget is $b_t=H-t$. This cap is
set by the training protocol; $b_t$ measures the remaining execution allowance
and does not estimate the number of steps needed to reach success.

For a task instruction $x$, we freeze the current trainable policy as
$\pi_{\mathrm{old}}$ at the beginning of each update and sample $G$ independent
interaction trajectories. Let $\tau_i$ be the $i$-th trajectory and $R_i$ its
final environment reward. The rollout group is
\begin{equation}
    \mathcal{G}_x=\{\tau_1,\ldots,\tau_G\},
    \qquad \tau_i\sim\pi_{\mathrm{old}}.
    \label{eq:root_group}
\end{equation}
The trainable policy is denoted by $\pi_\theta$. GRPO compares rewards within
$\mathcal{G}_x$ and assigns each trajectory a group-relative advantage $A_i$.
Reward variation therefore identifies which sampled behaviors should become
more or less likely. If every trajectory fails with the same reward, however,
\begin{equation}
    R_1=\cdots=R_G
    \quad\Longrightarrow\quad
    A_1=\cdots=A_G\approx 0.
    \label{eq:zero_signal}
\end{equation}
and the group provides little or no policy-gradient signal. We call this an
\emph{all-failure, zero-signal group}.

Groups with meaningful reward contrast follow the standard GRPO update.
VH-GPO intervenes only when every trajectory fails, the rewards
are equal up to a small tolerance, and the resulting advantages are numerically
zero. These groups define the failure cases addressed by our method.

\subsection{VH-GPO Overview}
\label{sec:method_overview}

\begin{figure*}[t]
    \centering
    \includegraphics[width=\textwidth]{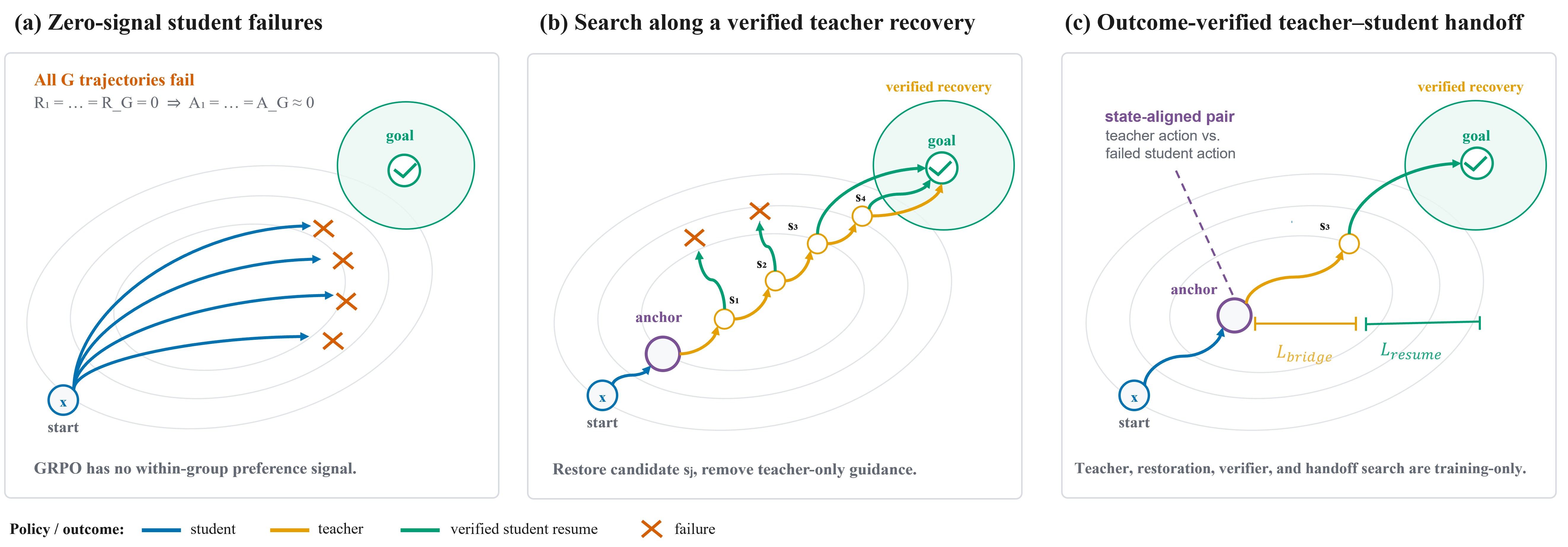}
    \caption{Overview of VH-GPO. (a) When every
    student trajectory fails, group-relative advantages provide no useful
    within-group preference. (b) A frozen teacher recovers from a restorable
    student state; candidate states along this verified recovery are restored
    and tested by the frozen student. (c) The
    earliest state from which student success is independently confirmed
    defines the handoff frontier. The retained trajectory consists of the
    original student prefix, a short teacher bridge, and a verified student
    resume; a state-aligned teacher--student divergence supplies the pairwise
    target.}
    \label{fig:method_overview}
\end{figure*}

VH-GPO constructs a verified training relay from an all-failure, zero-signal
group (Figure~\ref{fig:method_overview}). A frozen teacher first recovers from
a restorable student state. Within a bounded search budget, we then identify
the earliest tested state from which the frozen student's success is
independently confirmed. The relay retains the original student prefix,
the teacher bridge before handoff, and the verified student continuation:
\begin{equation}
    \boxed{
    \text{student prefix}
    \;\longrightarrow\;
    \text{teacher bridge}
    \;\longrightarrow\;
    \text{student resume}
    }.
    \label{eq:relay_overview}
\end{equation}

Only candidates that pass state restoration, teacher recovery, independent
student confirmation, and full-trajectory replay contribute auxiliary targets.

\subsection{Verified Teacher--Student Handoff and Relay Construction}
\label{sec:relay_construction}

\noindent\textbf{Obtaining a verified recovery.}

For each selected all-failure, zero-signal group, we choose one failed student
trajectory as the recovery \emph{backbone}. To reserve interaction budget for
recovery, we retain collected candidate states with $b_t\geq b_{\min}$, where
$b_{\min}$ is a configured minimum budget. We select the failed trajectory with
the deepest candidate satisfying the budget condition. An unsuccessful episode
may still contain useful task progress. By first attempting recovery from a
late student-visited state, we aim to build on this progress and obtain
corrective supervision for the situation created by the student's own actions,
reducing the need for the teacher to demonstrate earlier portions of the task
again. The group contributes no
auxiliary data if no eligible trajectory exists or the selected backbone lacks
an initial-state candidate.

We call the state where the frozen teacher $\pi_T$ begins acting the
\emph{anchor}. Starting from the deepest eligible candidate on the selected
backbone, we construct a bounded schedule of progressively earlier candidates
using a fixed backtracking stride, including the initial state as the final
fallback. At each anchor, we restore and verify the state, then let the teacher
attempt recovery using the task and visible interaction history within the
remaining interaction budget. We retain the first anchor yielding a verified
successful recovery; if all scheduled anchors fail, we discard the group.

If multiple attempts from the same anchor succeed, we keep the shorter recovery.
A retained recovery with $L$ teacher actions is written as
\begin{equation}
    \tau_T=(a^T_1,a^T_2,\ldots,a^T_L),
    \label{eq:teacher_recovery}
\end{equation}
where the superscript $T$ marks teacher actions.

\medskip
\noindent\textbf{Handoff selection and relay verification.}

The verified recovery shows that the task can be completed from the anchor, but
the teacher may not need to execute all $L$ actions. Let $s_j$ be the state
after its first $j$ actions, with $s_0$ denoting the anchor and $s_L$ the
terminal state. At a tested nonterminal state $s_j$, we return control to the
frozen rollout policy $\pi_{\mathrm{old}}$ and let it attempt the remainder of
the task without teacher assistance, subject to the original episode's
remaining interaction budget.

Within a fixed search budget, we test candidate handoff states and select the
earliest one with a successful student continuation. We then independently
validate this candidate using newly sampled continuations from the same state.
Search details are given in Appendix~\ref{sec:training_procedure}.
Let $\mathcal{J}_{\mathrm{test}}$ be the set of indices examined within this
search budget. We select
\begin{equation}
    j^\star
    =
    \min\left\{
        j\in\mathcal{J}_{\mathrm{test}}:
        \text{success from }s_j\text{ is independently confirmed}
    \right\}.
    \label{eq:minimal_handoff}
\end{equation}
We require $j^\star<L$ so that control returns before the teacher completes the
task.

The final relay concatenates the original student trajectory before the anchor,
the teacher actions $(a^T_1,\ldots,a^T_{j^\star})$, and the independently
confirmed student continuation from $s_{j^\star}$.

\subsection{Relay-based Training Objectives}
\label{sec:handoff_objectives}

Given a verified relay, we derive up to three complementary signals: an action
preference, teacher-bridge completion, and student-resume completion. The three
objectives are normalized independently. We write $s_\theta(y\mid h)$ for the
student's mean token log probability of target text $y$ given student-visible
context $h$.

\medskip
\noindent\textbf{Action Preference Objective.}

Let $h_d$ be the earliest verified prompt at which the teacher bridge and the
failed student trajectory choose different executable actions. At this prompt,
$a_{\mathrm{chosen}}$ is the teacher action from the successful relay and
$a_{\mathrm{rejected}}$ is the student action from the failed trajectory.
The current student's chosen-over-rejected score, computed over executable
action tokens, is
\begin{equation}
    p_\theta
    =
    s_\theta(a_{\mathrm{chosen}}\mid h_d)
    -
    s_\theta(a_{\mathrm{rejected}}\mid h_d).
    \label{eq:policy_preference}
\end{equation}
Let $p_{\mathrm{ref}}$ be the same score computed by a fixed reference policy
initialized from the student's starting checkpoint. Let $\sigma$ be the
logistic sigmoid and let the positive coefficient $\beta$ control preference
strength. We adapt the reference-anchored preference objective of
DPO~\citep{rafailov2023dpo} to state-aligned executable actions using the
mean-token scores above. The loss for one action pair is
\begin{equation}
    \ell_{\mathrm{pair}}
    =
    -\log\sigma\!\left(\beta(p_\theta-p_{\mathrm{ref}})\right).
    \label{eq:pair_loss}
\end{equation}
We denote the average over valid pairs by
$\mathcal{L}_{\mathrm{pair}}$. This objective teaches the student to prefer the
teacher action from a verified successful relay over the action observed in a
failed student trajectory at the same state, providing a localized correction
at their first action divergence.

\medskip
\noindent\textbf{Teacher Bridge Objective.}

Let $\mathcal{B}$ contain the prompt--response pairs from all verified teacher
bridges in the update. For each pair $(h,y)$, $h$ is the student-visible prompt
and $y$ is the complete teacher response for that turn. The teacher-bridge
objective is
\begin{equation}
    \mathcal{L}_{\mathrm{bridge}}
    =
    -\frac{1}{|\mathcal{B}|}
    \sum_{(h,y)\in\mathcal{B}}
    s_\theta(y\mid h).
    \label{eq:bridge_loss}
\end{equation}
This objective teaches the minimal corrective behavior needed to restore
student solvability. It is omitted when $\mathcal{B}$ is empty.

\medskip
\noindent\textbf{Student Resume Objective.}

Let $\mathcal{C}$ contain the prompt--response pairs from all independently
confirmed student resumes in the update, and let $|\mathcal{C}|$ be their
number. The student-resume objective is
\begin{equation}
    \mathcal{L}_{\mathrm{resume}}
    =
    -\frac{1}{|\mathcal{C}|}
    \sum_{(h,y)\in\mathcal{C}}
    s_\theta(y\mid h).
    \label{eq:resume_loss}
\end{equation}
This objective reinforces student-generated actions that are verified to lead
to success. It is an outcome-filtered positive completion objective rather than
an unbiased policy-gradient estimator, and it is omitted when $\mathcal{C}$ is
empty. It shares the motivation of learning from the agent's own good
experience with self-imitation learning~\citep{oh2018sil}. Unlike SIL's
off-policy actor--critic formulation, we use a positive completion loss on
independently verified student continuations following teacher intervention.

\medskip
\noindent\textbf{Overall Objective.}

Let the non-negative coefficients $\lambda_{\mathrm{pair}}$,
$\lambda_{\mathrm{bridge}}$, and $\lambda_{\mathrm{resume}}$ weight the three
relay-based losses. Their combined auxiliary objective is
\begin{equation}
    \mathcal{L}_{\mathrm{aux}}
    =
    \lambda_{\mathrm{pair}}\mathcal{L}_{\mathrm{pair}}
    +
    \lambda_{\mathrm{bridge}}\mathcal{L}_{\mathrm{bridge}}
    +
    \lambda_{\mathrm{resume}}\mathcal{L}_{\mathrm{resume}}.
    \label{eq:aux_loss}
\end{equation}
Let $\mathcal{L}_{\mathrm{GRPO}}$ denote the unchanged standard GRPO objective
and let $\mathcal{L}_{\mathrm{KL}}$ denote the reference-policy regularizer,
weighted by $\eta$. The optimizer combines the available terms as
\begin{equation}
    \mathcal{L}
    =
    \mathcal{L}_{\mathrm{GRPO}}
    +
    \eta\mathcal{L}_{\mathrm{KL}}
    +
    \mathcal{L}_{\mathrm{aux}}.
    \label{eq:total_loss}
\end{equation}

These terms are generally contributed by different rollout groups within the
same training batch. Groups with valid reward contrast contribute
$\mathcal{L}_{\mathrm{GRPO}}$. Zero-signal all-failure groups contribute
$\mathcal{L}_{\mathrm{aux}}$ only if recovery and relay verification succeed.
Thus, the total loss is combined at the batch or optimizer-update level; a
failed trajectory is not simultaneously treated as an ordinary GRPO positive
sample and a teacher-supervised positive target. The root GRPO and auxiliary
terms are normalized separately before combination. Concrete coefficient
values are specified in the experimental setup.

The complete training algorithm and deployment details are provided in
Appendix~\ref{sec:training_procedure}.

\section{Experiments}
\label{sec:experiments}

We evaluate VH-GPO with two questions in mind. First, does outcome-verified
teacher--student handoff improve the task performance of LLM agents?
Second, which parts of the method are responsible for the improvement?
We examine these questions on two long-horizon interactive benchmarks
at two model scales, complemented by controlled ablations and relay
diagnostics.

\subsection{Experimental Setup}
\label{sec:experimental_setup}

\paragraph{Benchmarks and metrics.}
We evaluate on ALFWorld~\citep{shridhar2021alfworld}, a text environment
with six embodied task families, and WebShop~\citep{yao2022webshop}, where
agents purchase products satisfying natural-language requests. We report
per-family and overall success on ALFWorld, and attribute-based partial-credit
score and exact success on WebShop. Main and ablation results are reported as
mean $\pm$ standard deviation in percent over three independent
runs with different random seeds. Reward definitions are given in
Appendix~\ref{sec:reward_design}.

\paragraph{Implementation Details.}
We train Qwen2.5-1.5B-Instruct and Qwen2.5-7B-Instruct
students~\citep{yang2024qwen25} from their instruction-tuned checkpoints for
150 updates using AdamW with learning rate $10^{-6}$, betas $(0.9,0.999)$,
and weight decay $0.01$. Each update uses 16 tasks with eight root rollouts
each. GRPO uses clipping parameter $0.2$; the KL coefficient is $\eta=0$
on ALFWorld and $\eta=0.01$ on WebShop. Interaction horizons for training
and evaluation are 50 steps on ALFWorld and 15 on WebShop.
Student responses are capped at 512 tokens.
We set the pairwise $\beta=0.1$ and
$(\lambda_{\mathrm{pair}},\lambda_{\mathrm{bridge}},
\lambda_{\mathrm{resume}})=(0.5,0.25,0.1)$, processing at most two eligible
zero-signal groups per update. The teacher is a frozen, GRPO-trained
Qwen2.5-1.5B-Instruct policy used only during training.
% Prompt templates are provided in Appendix~\ref{sec:prompt_templates}.

\paragraph{Baselines.}
We compare VH-GPO with representative and competitive RL training methods:
GRPO~\citep{shao2024deepseekmath}, GiGPO~\citep{feng2025gigpo},
GraphGPO~\citep{cheng2026graphgpo}, and HGPO~\citep{he2026hgpo}.
HGPO uses $K=4$ at both model scales.
Our reimplementation of OVCSD~\citep{xia2026ovcsd} and its results under
our experimental setting are reported in Appendix~\ref{sec:ovcsd_comparison}.

\subsection{Main Results}
\label{sec:main_results}

\begin{table*}[t]
    \centering
    \caption{Final performance on ALFWorld and WebShop (\%), reported as
    mean $\pm$ standard deviation over three seeds.
    Bold denotes the best mean within a model-size block.}
    \label{tab:main_results}
    \setlength{\tabcolsep}{2pt}
    \resizebox{\textwidth}{!}{%
    \begin{tabular}{llrrrrrrrrr}
        \toprule
        & & \multicolumn{7}{c}{ALFWorld} & \multicolumn{2}{c}{WebShop} \\
        \cmidrule(lr){3-9}\cmidrule(lr){10-11}
        Model & Method & Pick & Clean & Cool & Look & Heat & Pick2 & All
        & Score & Succ. \\
        \midrule
        1.5B & GRPO
        & \meanstd{75.00}{2.91} & \meanstd{74.73}{2.58} & \meanstd{63.06}{3.80} & \meanstd{50.00}{0.00}
        & \meanstd{38.46}{12.56} & \meanstd{35.01}{4.99} & \meanstd{59.38}{1.28} & \meanstd{85.98}{2.20} & \meanstd{71.62}{1.48} \\
        1.5B & GiGPO
        & \meanstd{94.05}{1.68} & \meanstd{\textbf{98.96}}{1.47} & \meanstd{69.12}{6.31} & \meanstd{77.78}{7.85}
        & \meanstd{\textbf{100.00}}{0.00} & \meanstd{92.64}{8.02} & \meanstd{90.63}{1.11} & \meanstd{84.83}{2.76} & \meanstd{71.09}{1.28} \\
        1.5B & GraphGPO
        & \meanstd{94.05}{3.37} & \meanstd{96.84}{0.05} & \meanstd{\textbf{84.56}}{2.55} & \meanstd{83.33}{0.00}
        & \meanstd{\textbf{100.00}}{0.00} & \meanstd{91.58}{1.62} & \meanstd{92.71}{0.74} & \meanstd{86.55}{0.97} & \meanstd{75.52}{3.21} \\
        1.5B & HGPO ($K=4$)
        & \meanstd{97.62}{1.68} & \meanstd{93.71}{2.51} & \meanstd{76.91}{0.51} & \meanstd{77.78}{7.85}
        & \meanstd{94.87}{3.63} & \meanstd{74.69}{0.44} & \meanstd{86.98}{0.97} & \meanstd{88.23}{1.13} & \meanstd{75.52}{2.42} \\
        1.5B & VH-GPO
        & \meanstd{\textbf{98.81}}{1.68} & \meanstd{95.80}{1.45} & \meanstd{83.04}{2.37} & \meanstd{\textbf{94.44}}{7.86}
        & \meanstd{97.44}{3.63} & \meanstd{\textbf{97.58}}{1.71} & \meanstd{\textbf{94.79}}{0.97} & \meanstd{\textbf{88.71}}{1.45} & \meanstd{\textbf{77.34}}{3.37} \\
        \midrule
        7B & GRPO
        & \meanstd{82.14}{2.91} & \meanstd{84.21}{0.24} & \meanstd{61.54}{5.57} & \meanstd{77.78}{7.85}
        & \meanstd{89.75}{3.63} & \meanstd{74.74}{2.60} & \meanstd{78.12}{0.64} & \meanstd{83.35}{2.84} & \meanstd{71.61}{4.70} \\
        7B & GiGPO
        & \meanstd{96.43}{0.00} & \meanstd{88.44}{1.33} & \meanstd{86.14}{0.31} & \meanstd{\textbf{100.00}}{0.00}
        & \meanstd{\textbf{100.00}}{0.00} & \meanstd{\textbf{94.01}}{3.34} & \meanstd{92.71}{0.37} & \meanstd{86.00}{1.17} & \meanstd{75.78}{0.64} \\
        7B & GraphGPO
        & \meanstd{98.81}{1.68} & \meanstd{95.83}{2.95} & \meanstd{89.32}{4.19} & \meanstd{88.89}{7.86}
        & \meanstd{94.87}{3.63} & \meanstd{93.96}{1.75} & \meanstd{94.53}{0.64} & \meanstd{88.18}{1.62} & \meanstd{78.38}{3.15} \\
        7B & HGPO ($K=4$)
        & \meanstd{97.62}{1.68} & \meanstd{91.63}{3.85} & \meanstd{90.91}{7.42} & \meanstd{94.44}{7.86}
        & \meanstd{\textbf{100.00}}{0.00} & \meanstd{\textbf{94.01}}{3.34} & \meanstd{94.27}{1.33} & \meanstd{88.59}{0.68} & \meanstd{78.12}{3.37} \\
        7B & VH-GPO
        & \meanstd{\textbf{100.00}}{0.00} & \meanstd{\textbf{97.88}}{1.50} & \meanstd{\textbf{98.48}}{2.14} & \meanstd{\textbf{100.00}}{0.00}
        & \meanstd{97.44}{3.63} & \meanstd{84.26}{6.34} & \meanstd{\textbf{95.57}}{0.98} & \meanstd{\textbf{88.81}}{0.56} & \meanstd{\textbf{79.69}}{2.92} \\
        \bottomrule
    \end{tabular}}
\end{table*}

Table~\ref{tab:main_results} shows that VH-GPO achieves the highest overall
ALFWorld success and both WebShop metrics among the compared methods at
both model scales. Compared with credit-assignment baselines, VH-GPO
addresses not only where to assign credit but also what successful behavior
is available to learn. For all-failure groups, it constructs a verified
alternative from a student-visited state, turning otherwise uninformative
outcomes into corrective action preferences and successful completion targets.

The 7B results further suggest that useful assistance need not come from a
larger teacher. Verification allows VH-GPO to select locally effective
corrections from a frozen 1.5B teacher while leaving the remaining task to
the 7B student. This division of labor exploits teacher recoveries without
replacing the student's own completion behavior as a training target.

\subsection{Training Dynamics}
\label{sec:training_dynamics}

\begin{figure}[t]
    \centering
    \includegraphics[width=\textwidth]{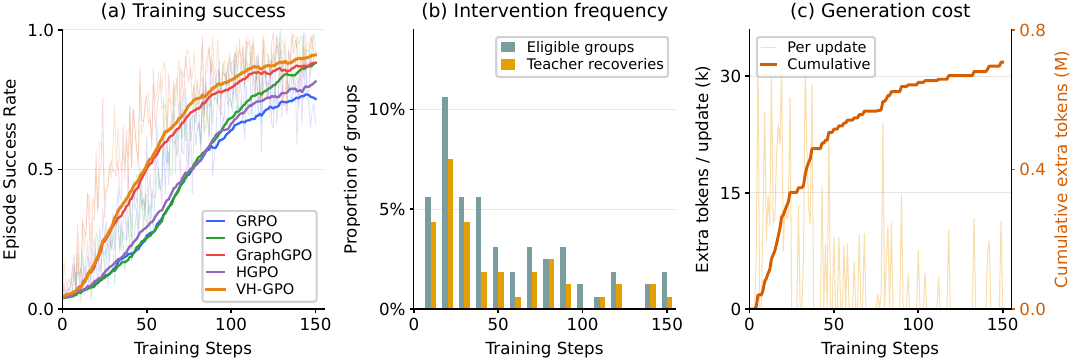}
    \caption{Training dynamics on ALFWorld with Qwen2.5-1.5B-Instruct.
    (a) Training episode success rate: light curves show per-update values;
    dark curves show EMA trends with decay $\alpha=0.95$.
    (b) VH-GPO intervention frequency: eligible zero-signal groups selected for
    recovery and verified teacher recoveries, as proportions of all training
    groups in ten-update windows (16 groups per update).
    (c) VH-GPO additional generation: per-update tokens (left axis, thousands)
    and cumulative tokens (right axis, millions), including teacher recovery,
    student search, confirmation, and subsequently discarded attempts.}
    \label{fig:alfworld_dynamics}
\end{figure}

To complement the final evaluation in Table~\ref{tab:main_results},
Figure~\ref{fig:alfworld_dynamics}(a) visualizes the training dynamics
on ALFWorld for the 1.5B student. VH-GPO enters the leading group early and
maintains a higher smoothed training success rate through most of the run.
Panels (b)--(c) show that selected zero-signal groups become less frequent
and cumulative additional output tokens grow more slowly later in training.

\subsection{Ablation Study}
\label{sec:ablations}

We ablate VH-GPO on ALFWorld with the 1.5B student using the same 150-update
budget for all variants. We separately examine the training objectives,
recovery and handoff mechanisms, and additional student rollout budget.
All ablation tables report success as mean $\pm$ standard deviation in
percent and the mean change $\Delta$ in percentage points relative to full
VH-GPO. Detailed ablation protocols are provided in
Appendix~\ref{sec:ablation_protocols}.

\subsubsection{Training Objectives}
\label{sec:objective_ablations}

\begin{table}[htbp]
    \centering
    \caption{Training-objective ablations on ALFWorld with the 1.5B student.}
    \label{tab:objective_ablations}
    \begin{tabular}{lrr}
        \toprule
        Method & Success & $\Delta$ \\
        \midrule
        VH-GPO & \meanstd{\textbf{94.79}}{0.97} & -- \\
        \quad w/o action preference & \meanstd{88.28}{1.68} & $-6.51$ \\
        \quad w/o student resume loss & \meanstd{90.62}{0.64} & $-4.17$ \\
        \quad w/o teacher bridge loss & \meanstd{80.47}{1.69} & $-14.32$ \\
        \bottomrule
    \end{tabular}
\end{table}

The variants in Table~\ref{tab:objective_ablations} each remove one auxiliary
loss while retaining relay construction and the other two objectives.
Removing the teacher bridge loss yields the largest decrease, from
$94.79\%$ to $80.47\%$. Equal failure rewards in these zero-signal groups
cannot distinguish useful corrective behavior. The bridge loss converts a
verified recovery into stepwise supervision, increasing the likelihood of
teacher responses that lead toward a student-solvable state. It thus trains
the sequence of decisions needed to escape the observed failure. Removing
this loss eliminates direct imitation of that sequence; the performance drop
supports the value of explicitly learning how to recover.

Removing action preference reduces success by 6.51 percentage points.
Bridge imitation still rewards the teacher's responses, but no longer
explicitly contrasts the chosen action with the failed student's action at
the same state. The drop suggests that this localized contrast adds useful
decision-level supervision beyond imitating positive targets alone.

Removing the student resume loss reduces success by 4.17 points. Verified
student continuations are still collected, but are no longer reinforced as
completion targets. This suggests that being able to sample a successful
continuation is not equivalent to consolidating that behavior through
training: learning the correction and learning to finish independently
provide complementary benefits.

\subsubsection{Recovery and Handoff Mechanisms}
\label{sec:mechanism_ablations}

\begin{table}[htbp]
    \centering
    \caption{Recovery and handoff ablations on ALFWorld with the 1.5B student.}
    \label{tab:mechanism_ablations}
    \begin{tabular}{lrr}
        \toprule
        Method & Success & $\Delta$ \\
        \midrule
        VH-GPO & \meanstd{\textbf{94.79}}{0.97} & -- \\
        Random Failure Anchor & \meanstd{90.89}{0.74} & $-3.90$ \\
        Random Verified Handoff & \meanstd{90.10}{0.98} & $-4.69$ \\
        Terminal Teacher & \meanstd{86.98}{1.33} & $-7.81$ \\
        \bottomrule
    \end{tabular}
\end{table}

Table~\ref{tab:mechanism_ablations} examines where teacher intervention starts and ends.
\emph{Random Failure Anchor} shuffles the non-root anchor order on the same
selected backbone, retaining the candidate set, attempt limits, and root
fallback; success falls by 3.90 points. \emph{Random Verified Handoff} retains
the discovery search and budget but uniformly selects an alternative
discovery-success state after excluding the earliest one. Independent
confirmation and full-trajectory replay remain required, and success falls
by 4.69 points. \emph{Terminal Teacher} instead supervises the full successful
teacher suffix, retains valid action preference pairs, and omits handoff
search and student-resume supervision while still requiring full-trajectory
replay. Its 7.81-point drop supports combining a short teacher correction with
a verified student continuation rather than relying on full-suffix teacher
supervision. Together, these comparisons support the designed anchor order
and early handoff selection.

\subsubsection{Additional Rollout Budget}
\label{sec:rollout_ablations}

\begin{table}[htbp]
    \centering
    \caption{Additional-rollout comparison on ALFWorld with the 1.5B student.}
    \label{tab:rollout_ablations}
    \begin{tabular}{lrr}
        \toprule
        Method & Success & $\Delta$ \\
        \midrule
        VH-GPO & \meanstd{\textbf{94.79}}{0.97} & -- \\
        GRPO & \meanstd{59.38}{1.28} & $-35.41$ \\
        Extra-rollout GRPO & \meanstd{64.32}{2.88} & $-30.47$ \\
        \bottomrule
    \end{tabular}
\end{table}

\emph{Extra-rollout GRPO} generates four additional complete student
trajectories for each selected zero-signal group and trains them with ordinary
GRPO, without a teacher; the budget rationale is given in
Appendix~\ref{sec:ablation_protocols}. It improves success over GRPO by
4.94 percentage points to 64.32\%, but remains 30.47 points below VH-GPO
(Table~\ref{tab:rollout_ablations}), indicating that this additional-sampling
configuration does not reproduce VH-GPO's gains.

\subsection{Generation Overhead}
\label{sec:generation_overhead}

Table~\ref{tab:generation_overhead} reports additional output tokens summed
over all 150 updates of the 7B runs, normalized by the corresponding root
rollout output tokens. The count includes teacher recovery, student handoff
search, and independent confirmation, including failed attempts and
ultimately discarded candidates. With recovery limited to at most two
selected zero-signal groups per update, additional generation amounts to
1.77\% on ALFWorld and 2.61\% on WebShop, with teacher recovery accounting
for most of the overhead. Appendix~\ref{sec:handoff_analysis} reports the
corresponding recovery yields and retained-relay statistics.

\begin{table}[htbp]
    \centering
    \caption{Additional generation in the 7B training runs. Each entry is a
    percentage of root rollout output tokens, summed over 150 updates.
    Search and Confirmation denote student continuations.}
    \label{tab:generation_overhead}
    \begin{tabular}{lrrrr}
        \toprule
        Benchmark & Teacher recovery & Search & Confirmation & Total \\
        \midrule
        ALFWorld & 1.04 & 0.47 & 0.26 & 1.77 \\
        WebShop & 2.07 & 0.41 & 0.13 & 2.61 \\
        \bottomrule
    \end{tabular}
\end{table}

\section{Conclusions}
\label{sec:conclusion}

VH-GPO learns from all-failure, zero-signal groups through verified teacher
bridges and student continuations, returning control at the earliest
independently confirmed handoff found within a bounded search budget.
Experiments show improved agent performance with modest additional
output-token overhead and no teacher at inference. Ablations suggest
complementary benefits from local correction and student-generated completion.

\paragraph{Limitations.}
VH-GPO requires restorable environment states and verifiable task outcomes.
Its budget-limited handoff search incurs additional rollouts and may miss
earlier feasible handoffs. More efficient search strategies remain future work.

\bibliography{references}

@inproceedings{shridhar2021alfworld,
  title ={{ALFWorld: Aligning Text and Embodied Environments for Interactive Learning}},
  author={Mohit Shridhar and Xingdi Yuan and Marc-Alexandre C\^ot\'e and Yonatan Bisk and Adam Trischler and Matthew Hausknecht},
  booktitle = {Proceedings of the International Conference on Learning Representations (ICLR)},
  year = {2021}
}

@inproceedings{yao2022webshop,
  title     = {{WebShop}: Towards Scalable Real-World Web Interaction with Grounded Language Agents},
  author    = {Yao, Shunyu and Chen, Howard and Yang, John and Narasimhan, Karthik},
  booktitle = {Advances in Neural Information Processing Systems},
  year      = {2022}
}

@article{shao2024deepseekmath,
  title   = {{DeepSeekMath}: Pushing the Limits of Mathematical Reasoning in Open Language Models},
  author  = {Shao, Zhihong and Wang, Peiyi and Zhu, Qihao and Xu, Runxin and Song, Junxiao and Bi, Xiao and Zhang, Haowei and Zhang, Mingchuan and Li, Y. K. and Wu, Y. and Guo, Daya},
  journal = {arXiv preprint arXiv:2402.03300},
  year    = {2024}
}

@inproceedings{feng2025gigpo,
  title     = {Group-in-Group Policy Optimization for {LLM} Agent Training},
  author    = {Feng, Lang and Xue, Zhenghai and Liu, Tingcong and An, Bo},
  booktitle = {Advances in Neural Information Processing Systems},
  year      = {2025}
}

@inproceedings{cheng2026graphgpo,
  title     = {Beyond Trajectory-Level Attribution: Graph-Based Credit Assignment for Agentic Reinforcement Learning},
  author    = {Cheng, Xin and He, Shuo and Feng, Lang and Xu, HaiYang and Yan, Ming and Feng, Lei and An, Bo},
  booktitle = {International Conference on Machine Learning},
  year      = {2026}
}

@inproceedings{he2026hgpo,
  title={Hierarchy-of-groups policy optimization for long-horizon agentic tasks},
  author={Shuo He and Lang Feng and Qi Wei and Xin Cheng and Lei Feng and Bo An},
  booktitle={International Conference on Learning Representations},
  year={2026}
}

@article{yang2024qwen25,
    title   = {Qwen2.5 Technical Report}, 
    author  = {An Yang and Baosong Yang and Beichen Zhang and Binyuan Hui and Bo Zheng and Bowen Yu and Chengyuan Li and Dayiheng Liu and Fei Huang and Haoran Wei and Huan Lin and Jian Yang and Jianhong Tu and Jianwei Zhang and Jianxin Yang and Jiaxi Yang and Jingren Zhou and Junyang Lin and Kai Dang and Keming Lu and Keqin Bao and Kexin Yang and Le Yu and Mei Li and Mingfeng Xue and Pei Zhang and Qin Zhu and Rui Men and Runji Lin and Tianhao Li and Tingyu Xia and Xingzhang Ren and Xuancheng Ren and Yang Fan and Yang Su and Yichang Zhang and Yu Wan and Yuqiong Liu and Zeyu Cui and Zhenru Zhang and Zihan Qiu},
    journal = {arXiv preprint arXiv:2412.15115},
    year    = {2024}
}

@article{schulman2017ppo,
  title   = {Proximal Policy Optimization Algorithms},
  author  = {Schulman, John and Wolski, Filip and Dhariwal, Prafulla and Radford, Alec and Klimov, Oleg},
  journal = {arXiv preprint arXiv:1707.06347},
  year    = {2017}
}

@inproceedings{ahmadian2024back,
  title={Back to Basics: Revisiting {REINFORCE} Style Optimization for Learning from Human Feedback in {LLM}s},
  author={Ahmadian, Arash and Cremer, Chris and Gall{\'e}, Matthias and Fadaee, Marzieh and Kreutzer, Julia and Pietquin, Olivier and {\"U}st{\"u}n, Ahmet and Hooker, Sara},
  booktitle={Proceedings of the 62nd Annual Meeting of the Association for Computational Linguistics (Volume 1: Long Papers)},
  year={2024}
}

@inproceedings{zhou2024archer,
  title={{ArCHer}: Training Language Model Agents via Hierarchical Multi-Turn {RL}},
  author={Zhou, Yifei and Zanette, Andrea and Pan, Jiayi and Levine, Sergey and Kumar, Aviral},
  booktitle = 	 {Proceedings of the 41st International Conference on Machine Learning},
  year={2024}
}

@inproceedings{ross2011dagger,
  title     = {A Reduction of Imitation Learning and Structured Prediction to No-Regret Online Learning},
  author    = {Ross, St{\'e}phane and Gordon, Geoffrey and Bagnell, Drew},
  booktitle = {Proceedings of the Fourteenth International Conference on Artificial Intelligence and Statistics},
  year      = {2011}
}

@inproceedings{rafailov2023dpo,
  title     = {Direct Preference Optimization: Your Language Model is Secretly a Reward Model},
  author    = {Rafailov, Rafael and Sharma, Archit and Mitchell, Eric and Ermon, Stefano and Manning, Christopher D. and Finn, Chelsea},
  booktitle = {Advances in Neural Information Processing Systems},
  year      = {2023}
}

@inproceedings{agarwal2024opd,
  title     = {On-Policy Distillation of Language Models: Learning from Self-Generated Mistakes},
  author    = {Agarwal, Rishabh and Vieillard, Nino and Zhou, Yongchao and Stanczyk, Piotr and Ramos, Sabela and Geist, Matthieu and Bachem, Olivier},
  booktitle = {International Conference on Learning Representations},
  year      = {2024}
}

@article{xia2026ovcsd,
  title   = {From Scoring to Acting: Outcome-Verified Comparative Self-Distillation for {LLM} Agents},
  author  = {Xia, Xu and Piao, Jinghua and Yang, Min and Lan, Xiaochong and Chen, Jiaju and Li, Yong},
  journal = {arXiv preprint arXiv:2607.27937},
  year    = {2026}
}

@inproceedings{yao2023react,
  title     = {{ReAct}: Synergizing Reasoning and Acting in Language Models},
  author    = {Yao, Shunyu and Zhao, Jeffrey and Yu, Dian and Du, Nan and Shafran, Izhak and Narasimhan, Karthik and Cao, Yuan},
  booktitle = {International Conference on Learning Representations},
  year      = {2023}
}

@inproceedings{yu2025dapo,
  title     = {{DAPO}: An Open-Source {LLM} Reinforcement Learning System at Scale},
  author    = {Yu, Qiying and Zhang, Zheng and Zhu, Ruofei and Yuan, Yufeng and Zuo, Xiaochen and Yue, Yu and Dai, Weinan and Fan, Tiantian and Liu, Gaohong and Liu, Juncai and Liu, Lingjun and Liu, Xin and Lin, Haibin and Lin, Zhiqi and Ma, Bole and Sheng, Guangming and Tong, Yuxuan and Zhang, Chi and Zhang, Mofan and Zhang, Ru and Zhang, Wang and Zhu, Hang and Zhu, Jinhua and Chen, Jiaze and Chen, Jiangjie and Wang, Chengyi and Yu, Hongli and Song, Yuxuan and Wei, Xiangpeng and Zhou, Hao and Liu, Jingjing and Ma, Wei-Ying and Zhang, Ya-Qin and Yan, Lin and Wu, Yonghui and Wang, Mingxuan},
  booktitle = {Advances in Neural Information Processing Systems},
  year      = {2025}
}

@article{wang2025ragen,
  title   = {{RAGEN}: Understanding Self-Evolution in {LLM} Agents via Multi-Turn Reinforcement Learning},
  author  = {Wang, Zihan and Wang, Kangrui and Wang, Qineng and Zhang, Pingyue and Li, Linjie and Yang, Zhengyuan and Jin, Xing and Yu, Kefan and Nguyen, Minh Nhat and Liu, Licheng and Gottlieb, Eli and Lu, Yiping and Cho, Kyunghyun and Wu, Jiajun and Li, Fei-Fei and Wang, Lijuan and Choi, Yejin and Li, Manling},
  journal = {arXiv preprint arXiv:2504.20073},
  year    = {2025}
}

@inproceedings{oh2018sil,
  title     = {Self-Imitation Learning},
  author    = {Oh, Junhyuk and Guo, Yijie and Singh, Satinder and Lee, Honglak},
  booktitle = {Proceedings of the 35th International Conference on Machine Learning},
  year      = {2018}
}

@inproceedings{kelly2019hgdagger,
  title     = {{HG-DAgger}: Interactive Imitation Learning with Human Experts},
  author    = {Kelly, Michael and Sidrane, Chelsea and Driggs-Campbell, Katherine and Kochenderfer, Mykel J.},
  booktitle = {IEEE International Conference on Robotics and Automation},
  year      = {2019}
}

@article{ouyang2022instructgpt,
  title = {{Training language models to follow instructions with human feedback}},
  author = {Ouyang, Long and Wu, Jeff and Jiang, Xu and Almeida, Diogo and Wainwright, Carroll L. and Mishkin, Pamela and Zhang, Chong and Agarwal, Sandhini and Slama, Katarina and Ray, Alex and Schulman, John and Hilton, Jacob and Kelton, Fraser and Miller, Luke and Simens, Maddie and Askell, Amanda and Welinder, Peter and Christiano, Paul and Leike, Jan and Lowe, Ryan},
  journal = {Advances in Neural Information Processing Systems},
  year = {2022}
}

@article{deepseekai2025r1,
   title={{DeepSeek-R1} incentivizes reasoning in {LLMs} through reinforcement learning},
   journal={Nature},
   publisher={Springer Science and Business Media LLC},
   author={Guo, Daya and Yang, Dejian and Zhang, Haowei and others},
   year={2025}
}

@article{nakano2022webgpt,
  title = {{WebGPT: Browser-assisted question-answering with human feedback}},
  author = {Reiichiro Nakano and Jacob Hilton and Suchir Balaji and Jeff Wu and Long Ouyang and Christina Kim and Christopher Hesse and Shantanu Jain and Vineet Kosaraju and William Saunders and Xu Jiang and Karl Cobbe and Tyna Eloundou and Gretchen Krueger and Kevin Button and Matthew Knight and Benjamin Chess and John Schulman},
  journal = {arXiv preprint arXiv:2112.09332},
  year = {2022}
}

@article{zheng2025gspo,
  title = {{Group Sequence Policy Optimization}},
  author = {Chujie Zheng and Shixuan Liu and Mingze Li and Xiong-Hui Chen and Bowen Yu and Chang Gao and Kai Dang and Yuqiong Liu and Rui Men and An Yang and Jingren Zhou and Junyang Lin},
  journal = {arXiv preprint arXiv:2507.18071},
  year = {2025}
}

@inproceedings{liu2025drgrpo,
  title = {Understanding {R1-Zero}-Like Training: A Critical Perspective},
  author = {Zichen Liu and Changyu Chen and Wenjun Li and Penghui Qi and Tianyu Pang and Chao Du and Wee Sun Lee and Min Lin},
  booktitle = {Conference on Language Modeling (COLM)},
  year = {2025}
}

@article{hu2025reinforcepp,
  title = {{REINFORCE++: Stabilizing Critic-Free Policy Optimization with Global Advantage Normalization}},
  author = {Jian Hu and Jason Klein Liu and Haotian Xu and Wei Shen},
  journal = {arXiv preprint arXiv:2501.03262},
  year = {2025}
}

@inproceedings{qi2025webrl,
  title = {{WebRL: Training LLM Web Agents via Self-Evolving Online Curriculum Reinforcement Learning}},
  author={Zehan Qi and Xiao Liu and Iat Long Iong and Hanyu Lai and Xueqiao Sun and Jiadai Sun and Xinyue Yang and Yu Yang and Shuntian Yao and Wei Xu and Jie Tang and Yuxiao Dong},
  booktitle = {International Conference on Learning Representations},
  year = {2025}
}

@inproceedings{bai2024digirl,
  title = {{DigiRL: Training In-The-Wild Device-Control Agents with Autonomous Reinforcement Learning}},
  author = {Hao Bai and Yifei Zhou and Mert Cemri and Jiayi Pan and Alane Suhr and Sergey Levine and Aviral Kumar},
  booktitle = {Advances in Neural Information Processing Systems},
  year = {2024}
}

@article{putta2024agentq,
  title = {{Agent Q: Advanced Reasoning and Learning for Autonomous AI Agents}},
  author = {Pranav Putta and Edmund Mills and Naman Garg and Sumeet Motwani and Chelsea Finn and Divyansh Garg and Rafael Rafailov},
  journal = {arXiv preprint arXiv:2408.07199},
  year = {2024}
}

@inproceedings{xi2024agentgym,
    title = "{A}gent{G}ym: Evaluating and Training Large Language Model-based Agents across Diverse Environments",
    author = "Xi, Zhiheng  and
      Ding, Yiwen  and
      Chen, Wenxiang  and
      Hong, Boyang  and
      Guo, Honglin  and
      Wang, Junzhe  and
      Guo, Xin  and
      Yang, Dingwen  and
      Liao, Chenyang  and
      He, Wei  and
      Gao, Songyang  and
      Chen, Lu  and
      Zheng, Rui  and
      Zou, Yicheng  and
      Gui, Tao  and
      Zhang, Qi  and
      Qiu, Xipeng  and
      Huang, Xuanjing  and
      Wu, Zuxuan  and
      Jiang, Yu-Gang",
    booktitle = "Proceedings of the 63rd Annual Meeting of the Association for Computational Linguistics (Volume 1: Long Papers)",
    year = "2025",
}

@inproceedings{xi2025agentgymrl,
  title = {{AgentGym-RL}: An Open-Source Framework to Train {LLM} Agents for Long-Horizon Decision Making via Multi-Turn {RL}},
  author={Zhiheng Xi and Jixuan Huang and Chenyang Liao and Baodai Huang and Jiaqi Liu and Honglin Guo and yajie yang and Rui Zheng and Junjie Ye and Jiazheng Zhang and Wenxiang Chen and Wei He and Yiwen Ding and Guanyu Li and Zehui Chen and Zhengyin Du and Xuesong Yao and Yufei Xu and Jiecao Chen and Tao Gui and Zuxuan Wu and Qi Zhang and Xuanjing Huang and Yu-Gang Jiang},
  booktitle = {International Conference on Learning Representations},
  year      = {2026}
}

@article{luo2025agentlightning,
  title = {{Agent Lightning: Train ANY AI Agents with Reinforcement Learning}},
  author = {Xufang Luo and Yuge Zhang and Zhiyuan He and Zilong Wang and Siyun Zhao and Dongsheng Li and Luna K. Qiu and Yuqing Yang},
  journal = {arXiv preprint arXiv:2508.03680},
  year = {2025}
}

@inproceedings{jin2025searchr1,
  title={Search-R1: Training {LLM}s to Reason and Leverage Search Engines with Reinforcement Learning},
  author={Bowen Jin and Hansi Zeng and Zhenrui Yue and Jinsung Yoon and Sercan O Arik and Dong Wang and Hamed Zamani and Jiawei Han},
  booktitle={Second Conference on Language Modeling},
  year={2025}
}

@inproceedings{kazemnejad2025vineppo,
  title = {{VinePPO: Refining Credit Assignment in RL Training of LLMs}},
  author = {Amirhossein Kazemnejad and Milad Aghajohari and Eva Portelance and Alessandro Sordoni and Siva Reddy and Aaron Courville and Nicolas Le Roux},
  booktitle={Forty-second International Conference on Machine Learning},
  year = {2025}
}

@inproceedings{lightman2023verify,
  title = {{Let's Verify Step by Step}},
  author = {Lightman, Hunter and Kosaraju, Vineet and Burda, Yuri and Edwards, Harrison and Baker, Bowen and Lee, Teddy and Leike, Jan and Schulman, John  and Sutskever, Ilya and Cobbe, Karl},
  booktitle = {International Conference on Learning Representations},
  year = {2024}
}

@inproceedings{wang2024mathshepherd,
  title = {{Math-Shepherd: Verify and Reinforce LLMs Step-by-step without Human Annotations}},
  author = {Peiyi Wang and Lei Li and Zhihong Shao and R. X. Xu and Damai Dai and Yifei Li and Deli Chen and Y. Wu and Zhifang Sui},
  booktitle = {Proceedings of the 62nd Annual Meeting of the Association for Computational Linguistics (Volume 1: Long Papers)},
  year = {2024}
}

@inproceedings{xie2024mctsdpo,
  title = {{Monte Carlo Tree Search Boosts Reasoning via Iterative Preference Learning}},
  author = {Yuxi Xie and Anirudh Goyal and Wenyue Zheng and Min-Yen Kan and Timothy P. Lillicrap and Kenji Kawaguchi and Michael Shieh},
  booktitle = {The First Workshop on System-2 Reasoning at Scale, NeurIPS'24},
  year = {2024}
}

@inproceedings{zhang2024restmcts,
  title = {{ReST-MCTS*: LLM Self-Training via Process Reward Guided Tree Search}},
  author = {Zhang, Dan and Zhoubian, Sining and Hu, Ziniu and Yue, Yisong and Dong, Yuxiao and Tang, Jie},
  booktitle = {Advances in Neural Information Processing Systems},
  year = {2024}
}

@inproceedings{zelikman2022star,
  title = {{STaR: Bootstrapping Reasoning With Reasoning}},
  author = {Zelikman, Eric and Wu, Yuhuai and Mu, Jesse and Goodman, Noah},
  booktitle = {Advances in Neural Information Processing Systems},
  year = {2022}
}

@article{gulcehre2023rest,
  title = {{Reinforced Self-Training (ReST) for Language Modeling}},
  author = {Caglar Gulcehre and Tom Le Paine and Srivatsan Srinivasan and Ksenia Konyushkova and Lotte Weerts and Abhishek Sharma and Aditya Siddhant and Alex Ahern and Miaosen Wang and Chenjie Gu and Wolfgang Macherey and Arnaud Doucet and Orhan Firat and Nando de Freitas},
  journal = {arXiv preprint arXiv:2308.08998},
  year = {2023},
}

@article{singh2024restem,
  title = {{Beyond Human Data: Scaling Self-Training for Problem-Solving with Language Models}},
  author={Avi Singh and John D Co-Reyes and Rishabh Agarwal and Ankesh Anand and Piyush Patil and Xavier Garcia and Peter J Liu and James Harrison and Jaehoon Lee and Kelvin Xu and Aaron T Parisi and Abhishek Kumar and Alexander A Alemi and Alex Rizkowsky and Azade Nova and Ben Adlam and Bernd Bohnet and Gamaleldin Fathy Elsayed and Hanie Sedghi and Igor Mordatch and Isabelle Simpson and Izzeddin Gur and Jasper Snoek and Jeffrey Pennington and Jiri Hron and Kathleen Kenealy and Kevin Swersky and Kshiteej Mahajan and Laura A Culp and Lechao Xiao and Maxwell Bileschi and Noah Constant and Roman Novak and Rosanne Liu and Tris Warkentin and Yamini Bansal and Ethan Dyer and Behnam Neyshabur and Jascha Sohl-Dickstein and Noah Fiedel},
  journal = {Transactions on Machine Learning Research},
  year = {2024}
}

@article{dong2023raft,
  title={{RAFT}: Reward rAnked FineTuning for Generative Foundation Model Alignment},
  author={Hanze Dong and Wei Xiong and Deepanshu Goyal and Yihan Zhang and Winnie Chow and Rui Pan and Shizhe Diao and Jipeng Zhang and KaShun SHUM and Tong Zhang},
  journal={Transactions on Machine Learning Research},
  year={2023}
}

@article{aksitov2023restreact,
  title = {{ReST meets ReAct: Self-Improvement for Multi-Step Reasoning LLM Agent}},
  author = {Renat Aksitov and Sobhan Miryoosefi and Zonglin Li and Daliang Li and Sheila Babayan and Kavya Kopparapu and Zachary Fisher and Ruiqi Guo and Sushant Prakash and Pranesh Srinivasan and Manzil Zaheer and Felix Yu and Sanjiv Kumar},
  journal = {arXiv preprint arXiv:2312.10003},
  year = {2023}
}

@inproceedings{gu2024minillm,
  title = {{MiniLLM: Knowledge Distillation of Large Language Models}},
  author = {Yuxian Gu and Li Dong and Furu Wei and Minlie Huang},
  booktitle = {International Conference on Learning Representations},
  year = {2024}
}

@inproceedings{hsieh2023distilling,
  title     = {Distilling Step-by-Step! Outperforming Larger Language Models with Less Training Data and Smaller Model Sizes},
  author    = {Hsieh, Cheng-Yu and Li, Chun-Liang and Yeh, Chih-kuan and Nakhost, Hootan and Fujii, Yasuhisa and Ratner, Alex and Krishna, Ranjay and Lee, Chen-Yu and Pfister, Tomas},
  booktitle = {Findings of the Association for Computational Linguistics: ACL 2023},
  year      = {2023}
}

@inproceedings{hoque2021thriftydagger,
  title = {{ThriftyDAgger: Budget-Aware Novelty and Risk Gating for Interactive Imitation Learning}},
  author =  {Hoque, Ryan and Balakrishna, Ashwin and Novoseller, Ellen and Wilcox, Albert and Brown, Daniel S. and Goldberg, Ken},
  booktitle = 	 {Proceedings of the 5th Conference on Robot Learning},
  year = {2022}
}

@inproceedings{zhang2016safedagger,
  title = {{Query-Efficient Imitation Learning for End-to-End Autonomous Driving}},
  author = {Jiakai Zhang and Kyunghyun Cho},
  booktitle = {Proceedings of the Thirty-First AAAI Conference on Artificial Intelligence},
  year = {2017}
}

@inproceedings{uchendu2023jsrl,
  title = {{Jump-Start Reinforcement Learning}},
  author = {Ikechukwu Uchendu and Ted Xiao and Yao Lu and Banghua Zhu and Mengyuan Yan and Jos{\'e}phine Simon and Matthew Bennice and Chuyuan Fu and Cong Ma and Jiantao Jiao and Sergey Levine and Karol Hausman},
  booktitle = {Proceedings of the 40th International Conference on Machine Learning},
  year = {2023}
}

@article{ecoffet2021goexplore,
  title = {{First return, then explore}},
  author = {Adrien Ecoffet and Joost Huizinga and Joel Lehman and Kenneth O. Stanley and Jeff Clune},
  journal = {Nature},
  year = {2021}
}

@article{ji2026actguide,
  title = {{Learning Agentic Policy from Action Guidance}},
  author = {Yuxiang Ji and Zengbin Wang and Yong Wang and Shidong Yang and Ziyu Ma and Guanhua Chen and Zonghua Sun and Liaoni Wu and Xiangxiang Chu},
  journal = {arXiv preprint arXiv:2605.12004},
  year = {2026}
}

@article{xia2026dashopd,
  title = {{DASH-OPD: Discrepancy-Aware Switching with Hysteresis for On-Policy Distillation}},
  author = {Yuchen Xia and Qianguo Sun and Chao Song and Junlong Wu and Yiyan Qi and Yunjian Xu},
  journal = {arXiv preprint arXiv:2607.29078},
  year = {2026}
}

@inproceedings{shinn2023reflexion,
  title = {{Reflexion: Language Agents with Verbal Reinforcement Learning}},
  author = {Shinn, Noah and Cassano, Federico and Gopinath, Ashwin and Narasimhan, Karthik and Yao, Shunyu},
  booktitle = {Advances in Neural Information Processing Systems},
  year = {2023}
}
\bibliographystyle{iclr2027_conference}

% Appendices must appear after the references.
\clearpage
\appendix
\raggedbottom
\section{Training Algorithm and Deployment}
\label{sec:training_procedure}

\begin{center}
\begin{minipage}{0.96\linewidth}
\hrule
\vspace{0.5em}
\textbf{Algorithm 1: VH-GPO training}
\vspace{0.3em}
\begin{tabbing}
\hspace{1.5em}\=\hspace{1.5em}\=\hspace{1.5em}\=\kill
\textbf{Input:} task batch $\mathcal{X}$, student $\pi_\theta$, frozen teacher
$\pi_T$, intervention cap $M$ \\
\textbf{Output:} updated student parameters $\theta$ \\
Freeze the current student as $\pi_{\mathrm{old}}$ \\
Initialize the GRPO and auxiliary loss accumulators \\
Initialize the number of selected zero-signal groups $m\leftarrow 0$ \\
\textbf{for each} task $x\in\mathcal{X}$ \textbf{do} \\
\>Collect rollout group $\mathcal{G}_x$ using $\pi_{\mathrm{old}}$ \\
\>\textbf{if} $\mathcal{G}_x$ has useful reward contrast \textbf{then} \\
\>\>Accumulate its standard GRPO objective \\
\>\>\textbf{continue} \\
\>\textbf{end if} \\
\>\textbf{if} $\mathcal{G}_x$ is not an all-failure, zero-signal group
\textbf{then continue} \\
\>\textbf{if} $m=M$ \textbf{then continue}; otherwise set $m\leftarrow m+1$ \\
\>Select a backbone by the deepest candidate satisfying the budget condition \\
\>\textbf{if} the backbone or its initial-state candidate is missing
\textbf{then continue} \\
\>Schedule anchors on this backbone from deep to shallow, ending at the root \\
\>Mark the teacher recovery as unavailable \\
\>\textbf{for each} scheduled anchor \textbf{do} \\
\>\>Restore the anchor and verify the reconstructed state \\
\>\>\textbf{if} restoration fails \textbf{then continue} \\
\>\>Sample teacher recovery attempts from the restored anchor \\
\>\>\textbf{if} the environment verifies any success \textbf{then} \\
\>\>\>Retain the shortest successful recovery and its anchor; \textbf{break} \\
\>\>\textbf{end if} \\
\>\textbf{end for} \\
\>\textbf{if} no teacher recovery was retained \textbf{then continue} \\
\>Find the earliest independently confirmed student handoff point \\
\>\textbf{if} no such point exists \textbf{then continue} \\
\>Build and replay the prefix--bridge--resume relay \\
\>\textbf{if} replay or information-leakage checks fail \textbf{then continue} \\
\>Accumulate its available pair, bridge, and resume objectives \\
\textbf{end for} \\
Normalize the GRPO and auxiliary objectives separately \\
Update $\theta$ using their combined objective and \textbf{return} $\theta$
\end{tabbing}
\vspace{0.3em}
\hrule
\end{minipage}
\end{center}

Any failure in restoration, teacher recovery, student confirmation, relay
replay, or auxiliary-example construction causes that candidate to contribute
no teacher-derived gradient; other groups continue through their normal paths.
At deployment, only the trained student $\pi_\theta(a_t\mid h_t)$ remains. The
teacher, state restoration, verifier, relay search, and handoff mechanism are
all removed. VH-GPO is therefore a training-only procedure and introduces no
teacher dependency at inference time.

\paragraph{Anchor selection and recovery budgets.}
Table~\ref{tab:anchor_parameters} lists the recovery-search settings used at
both student model scales. Starting from the deepest eligible anchor on the
selected backbone, each backtrack selects the deepest available candidate
whose depth is at least one stride earlier than the current anchor. The
schedule contains the initial anchor and at most four backtracked anchors.
If the root is not already included, it is appended as the final fallback,
giving at most six anchors per group. The actual number can be smaller when
fewer candidates are available. Anchor search stops at the first anchor
yielding a verified successful teacher recovery. Thus, the upper bound is
12 teacher recovery attempts per selected group; this bound excludes student
handoff-search and confirmation rollouts.

\begin{table}[htbp]
    \centering
    \caption{Anchor-selection and teacher-recovery parameters for both the
    1.5B and 7B students. Step counts refer to environment interactions.}
    \label{tab:anchor_parameters}
    \small
    \begin{tabular}{lrr}
        \toprule
        Parameter & ALFWorld & WebShop \\
        \midrule
        Minimum remaining budget $b_{\min}$ (steps) & 16 & 6 \\
        Anchor backtracking stride (steps) & 8 & 2 \\
        Maximum backtracks & 4 & 4 \\
        Maximum anchors per group, including root & 6 & 6 \\
        Teacher recovery attempts per anchor & 2 & 2 \\
        \bottomrule
    \end{tabular}
\end{table}

\paragraph{Handoff search.}
Discovery begins near the end of the verified teacher recovery and moves
backward with exponentially increasing jumps. After the first tested failure,
the search tests intermediate states in increasing order between that state
and the earliest later discovery-successful state, or the terminal state if
no discovery success has been found. This refinement stops at the first
success or when the search budget is exhausted. The earliest
discovery-successful state among all tested candidates is selected for
independent confirmation.

\section{Reward Design and Evaluation Metrics}
\label{sec:reward_design}

\paragraph{ALFWorld.}
The environment checks whether the final state satisfies all conditions of
the task, including the required object, its state, and its destination.
For the text-only setting used here, the task reward is
$r_{\mathrm{task}}=10\,\mathbf{1}[\mathrm{won}]$: successful episodes receive
10 and unsuccessful episodes receive 0. Intermediate task progress does not
provide a separate positive task reward. We report the percentage of
episodes with $\mathrm{won}=1$ for each task family and overall.

\paragraph{WebShop.}
The environment assigns a partial-credit score when the agent purchases a
product~\citep{yao2022webshop}. Let $N_{\mathrm{attr}}$ and
$N_{\mathrm{opt}}$ be the numbers of requested attributes and options,
$M_{\mathrm{attr}}$ and $M_{\mathrm{opt}}$ the numbers matched by the
purchased product, and $P$ indicate that its price meets the upper bound.
The original task score is
\begin{equation}
    s=r_{\mathrm{type}}\,
    \frac{M_{\mathrm{attr}}+M_{\mathrm{opt}}+P}
    {N_{\mathrm{attr}}+N_{\mathrm{opt}}+1},
\end{equation}
where $r_{\mathrm{type}}\in[0,1]$ measures whether the product type matches
the request. The score is zero before purchase and lies in $[0,1]$ at
purchase. Our training environment retains $s$ for evaluation but uses the
binary task reward
$r_{\mathrm{task}}=10\,\mathbf{1}[\mathrm{purchase\ completed}\ \land\ s=1]$.
Thus, partial credit does not become a positive task reward during training.
The reported WebShop \emph{Score} is $100$ times the mean $s$, whereas
\emph{Succ.} is $100$ times the fraction of episodes with $s=1$.

In both environments, the training configuration additionally subtracts
$0.1$ for each invalidly formatted action. This penalty is separate from
the task reward and is not included in the success-rate or WebShop Score
metrics above.

\section{Comparison with an OVCSD Reimplementation}
\label{sec:ovcsd_comparison}

An official implementation of OVCSD~\citep{xia2026ovcsd} was not publicly
available when we conducted these experiments. To compare the methods under
the same experimental setting, we reimplemented OVCSD following the
description in its paper and evaluated it using the setup in
Section~\ref{sec:experimental_setup}. Table~\ref{tab:ovcsd_comparison}
reports our reimplementation results, not the scores reported by the
original authors. Differences in implementation and experimental settings
may contribute to the gap from the original report; these results should
therefore be interpreted as a comparison under our evaluation protocol.

\begin{table}[htbp]
    \centering
    \caption{OVCSD reimplementation results under our experimental setting,
    reported as mean $\pm$ standard deviation over three seeds (\%).
    VH-GPO results from
    Table~\ref{tab:main_results} are repeated for reference. Bold denotes
    the best mean within each model-size block in this table.}
    \label{tab:ovcsd_comparison}
    \setlength{\tabcolsep}{2pt}
    \resizebox{\textwidth}{!}{%
    \begin{tabular}{llrrrrrrrrr}
        \toprule
        & & \multicolumn{7}{c}{ALFWorld} & \multicolumn{2}{c}{WebShop} \\
        \cmidrule(lr){3-9}\cmidrule(lr){10-11}
        Model & Method & Pick & Clean & Cool & Look & Heat & Pick2 & All
        & Score & Succ. \\
        \midrule
        1.5B & OVCSD
        & \meanstd{84.52}{1.68} & \meanstd{77.86}{4.62} & \meanstd{63.06}{3.80} & \meanstd{38.89}{7.86}
        & \meanstd{84.62}{0.00} & \meanstd{57.81}{4.66} & \meanstd{71.36}{1.33} & \meanstd{83.85}{2.61} & \meanstd{72.40}{1.33} \\
        1.5B & VH-GPO
        & \meanstd{\textbf{98.81}}{1.68} & \meanstd{\textbf{95.80}}{1.45} & \meanstd{\textbf{83.04}}{2.37} & \meanstd{\textbf{94.44}}{7.86}
        & \meanstd{\textbf{97.44}}{3.63} & \meanstd{\textbf{97.58}}{1.71} & \meanstd{\textbf{94.79}}{0.97} & \meanstd{\textbf{88.71}}{1.45} & \meanstd{\textbf{77.34}}{3.37} \\
        \midrule
        7B & OVCSD
        & \meanstd{\textbf{100.00}}{0.00} & \meanstd{\textbf{100.00}}{0.00} & \meanstd{70.85}{3.78} & \meanstd{77.78}{7.85}
        & \meanstd{94.87}{3.63} & \meanstd{77.12}{1.53} & \meanstd{88.54}{1.33} & \meanstd{82.91}{1.27} & \meanstd{76.04}{2.05} \\
        7B & VH-GPO
        & \meanstd{\textbf{100.00}}{0.00} & \meanstd{97.88}{1.50} & \meanstd{\textbf{98.48}}{2.14} & \meanstd{\textbf{100.00}}{0.00}
        & \meanstd{\textbf{97.44}}{3.63} & \meanstd{\textbf{84.26}}{6.34} & \meanstd{\textbf{95.57}}{0.98} & \meanstd{\textbf{88.81}}{0.56} & \meanstd{\textbf{79.69}}{2.92} \\
        \bottomrule
    \end{tabular}}
\end{table}

Under this protocol, VH-GPO achieves higher overall ALFWorld success and
both WebShop metrics than our OVCSD reimplementation at both scales.
Whereas OVCSD supervises verified teacher suffixes, VH-GPO concentrates
teacher imitation on a corrective bridge and reinforces the student's own
successful continuation. VH-GPO returns control after verifying that the
student can complete the remaining task independently. It limits teacher
imitation to the corrective actions before this handoff. These results
suggest that the value of teacher assistance depends not only on obtaining
a successful recovery, but also on using it to teach local correction while
reinforcing independent student completion.

\section{Ablation Protocols}
\label{sec:ablation_protocols}

\paragraph{Random Verified Handoff.}
This variant uses the same discovery search procedure and rollout budget as
full VH-GPO. Each tested nonterminal state receives two student continuations;
a state enters the discovery-success set $S$ if at least one continuation
succeeds. Discovery uses at most eight rollouts per teacher recovery. Let
$j_{\mathrm{early}}=\min S$ denote the earliest discovery-success index,
which full VH-GPO selects for independent confirmation. When alternative
discovery-success states are available, Random Verified Handoff samples
uniformly from $S\setminus\{j_{\mathrm{early}}\}$. Thus, the discovery
procedure, tested candidate set, and budget limit are unchanged, but the
final sampling set explicitly excludes the earliest successful candidate.
The selected state receives four additional confirmation continuations using
seeds distinct from discovery. At least one must succeed before the complete
student-prefix--teacher-bridge--student-resume action sequence is replayed
from the initial state. Only a successful replay qualifies the relay for
auxiliary training.

\paragraph{Random Failure Anchor.}
This variant first selects the same failed backbone as full VH-GPO: the
trajectory with the deepest budget-eligible candidate, with trajectory ID
used to break depth ties. It then constructs the same anchor list using the
remaining-interaction-budget constraint, backtracking stride, and limit on
anchor attempts. Only the non-root anchors are randomly permuted; the root
anchor remains last. After all teacher attempts at one anchor fail, the
procedure advances to the next anchor in this order and stops at the first
verified recovery. If all non-root anchors fail, it still attempts recovery
from the root. The candidate set and attempt limits are unchanged, but the
next anchor need not be shallower than the previous one. Actual resource use
depends on when the first recovery succeeds. Handoff search and confirmation
follow the full method.

\paragraph{Extra-rollout GRPO.}
We choose the number of extra student trajectories by expressing VH-GPO's
configured token reservation for confirmation continuations in units of
complete student rollouts. On ALFWorld, the theoretical maximum output-token
budget for one complete trajectory is
\begin{equation}
    B_{\mathrm{trajectory}} = 50 \times 512 = 25{,}600,
\end{equation}
using the environment-step cap and maximum response length. The configuration
reserves four such budgets, corresponding to the four parallel student
continuations used for independent confirmation:
\begin{equation}
    B_{\mathrm{reserve}} = 4 B_{\mathrm{trajectory}} = 102{,}400,
    \qquad
    N_{\mathrm{extra}}
    = \frac{B_{\mathrm{reserve}}}{B_{\mathrm{trajectory}}} = 4.
\end{equation}
Thus, four extra trajectories are chosen because their combined maximum
output-token budget equals the configured reservation for VH-GPO's four
parallel confirmation continuations. This conversion provides a
configuration-based budget rationale for allocating additional full student
rollouts to each selected zero-signal group. Training these rollouts with
ordinary GRPO tests whether spending this allocation on independent student
exploration can reproduce the gains from verified teacher recovery and
student resumption.

\section{Handoff Behavior}
\label{sec:handoff_analysis}

Table~\ref{tab:handoff_behavior} summarizes recovery and relay construction
during the 7B training runs. Eligible groups are all-failure groups that pass
the zero-signal checks and are selected for recovery attempts. A recovery is
counted when the environment verifies the teacher's terminal success. A
retained relay additionally requires independently confirmed nonterminal
student handoff, successful full-trajectory replay, and prompt-sanitization
checks. Successful replay of a full teacher recovery alone does not qualify
it for auxiliary training.

\begin{table}[ht]
    \centering
    \caption{Relay diagnostics for the 7B training runs. Parentheses report
    percentages relative to eligible groups for Recoveries and verified recoveries
    for Retained relays. Only retained
    relays contribute auxiliary training samples.}
    \label{tab:handoff_behavior}
    \small
    \begin{tabular}{lrrr}
        \toprule
        Benchmark & Eligible & Recoveries & Retained relays \\
        \midrule
        ALFWorld & 50 & 25 (50.0) & 18 (72.0) \\
        WebShop & 150 & 29 (19.3) & 10 (34.5) \\
        \bottomrule
    \end{tabular}
\end{table}

The teacher successfully recovers 25 of 50 selected groups on ALFWorld and
29 of 150 on WebShop. All these recoveries pass full-trajectory replay, but
only 18 and 10, respectively, also yield an independently confirmed
nonterminal handoff and are retained for auxiliary training. These represent
$72.0\%$ and $34.5\%$ of verified recoveries. The remaining 7 ALFWorld and
19 WebShop recoveries lack a confirmed handoff and contribute no
auxiliary training samples, despite successful replay of the teacher-completed
trajectory. This distinction shows that teacher success does not by itself
establish student resumability. Every retained relay in these runs also
provides a state-aligned action preference pair. No state-restoration failures
or privileged-information leaks are recorded in the selected groups.

The temporal distribution of these intervention events is shown in
Figure~\ref{fig:alfworld_dynamics}(b). Bars aggregate ten training
updates at a time for the 1.5B ALFWorld run. They report the proportion of all
training groups selected as eligible zero-signal groups and the proportion
that produce a verified teacher recovery; they are process diagnostics rather
than final benchmark metrics.

\subsection{Intervention Length}
\label{sec:intervention_length}

Sparse intervention and short intervention are distinct properties. To
measure the latter directly, we analyze the same 7B training runs over all
150 updates. For each retained relay, let $L$ be the number of actions in
the verified full teacher recovery from the selected anchor, $j^\star$ the
number retained in the teacher bridge, and $R$ the number in the selected
successful student resume suffix. Lengths count executed environment actions,
not tokens, and exclude the student prefix before the anchor. We recover $R$
from the successful independent-confirmation continuation selected by the
training procedure: shortest action length, with output tokens and seed as
tie-breakers. The retained teacher fraction is $j^\star/L$, and
$L-j^\star$ counts teacher actions omitted from bridge supervision.

\begin{figure}[!htb]
    \centering
    \includegraphics[width=\textwidth]{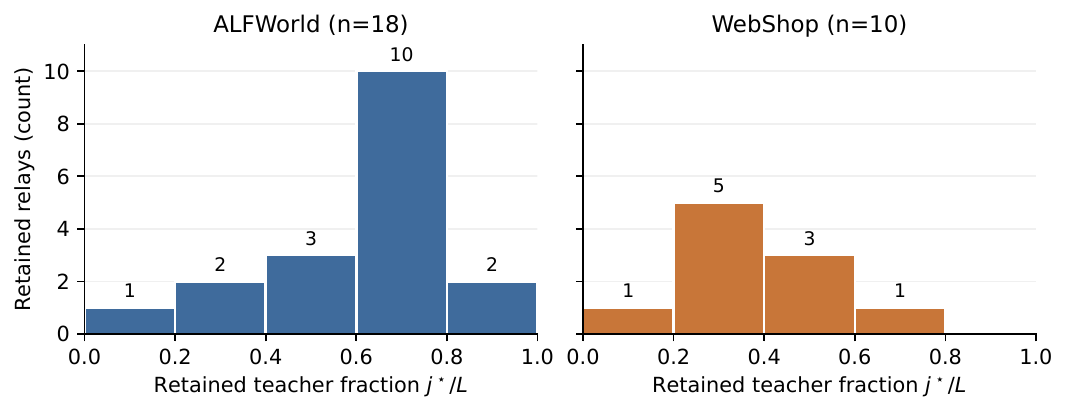}
    \caption{Distribution of the retained teacher fraction $j^\star/L$ for
    valid retained relays in the 7B runs. Each relay contributes one observation;
    bins have width 0.2 and include the left but not the right boundary
    (the final bin includes 1). Smaller values indicate earlier handoff
    relative to the verified teacher recovery.}
    \label{fig:intervention_lengths}
\end{figure}

Table~\ref{tab:intervention_lengths} and Figure~\ref{fig:intervention_lengths}
report these paired measurements for the 18 ALFWorld and 10 WebShop relays
in Table~\ref{tab:handoff_behavior}. Recoveries without a valid retained relay
are excluded, rather than treated as zero-length interventions. For context,
across all 25 and 29 verified recoveries, respectively, the mean full teacher
length is 13.44 and 4.48 actions (medians: 12 and 5).

\begin{table}[htbp]
    \centering
    \caption{Intervention lengths for retained relays in the 7B training runs.
    Paired entries are mean / median over relays; lengths are action counts.
    The pooled omitted fraction weights each relay by its full teacher length.}
    \label{tab:intervention_lengths}
    \begin{tabular}{lrr}
        \toprule
        Metric & ALFWorld & WebShop \\
        \midrule
        Retained relays & 18 & 10 \\
        Full teacher recovery $L$ & 14.78 / 14.00 & 5.20 / 5.00 \\
        Retained teacher bridge $j^\star$ & 9.39 / 9.50 & 1.80 / 2.00 \\
        Student resume suffix $R$ & 4.33 / 3.00 & 4.10 / 4.00 \\
        Retained teacher fraction $j^\star/L$ (\%) & 60.68 / 65.94 & 35.33 / 33.33 \\
        Omitted teacher actions $L-j^\star$ & 5.39 / 4.50 & 3.40 / 4.00 \\
        \midrule
        Total omitted / full teacher actions & 97 / 266 & 34 / 52 \\
        Pooled omitted fraction (\%) & 36.47 & 65.38 \\
        \bottomrule
    \end{tabular}
\end{table}

\FloatBarrier

Across retained relays, handoff removes 97 of 266 teacher actions on ALFWorld
and 34 of 52 on WebShop from bridge supervision, corresponding to pooled
reductions of 36.47\% and 65.38\%. The mean per-relay reductions,
$1-\operatorname{mean}(j^\star/L)$, are 39.32\% and 64.67\%, respectively.
Thus, teacher supervision is shortened within individual relays, beyond the
sparsity of intervention events. The distribution also shows that the bridge
is not uniformly short: its required length varies across tasks and benchmarks.
Here, \emph{minimal} means the earliest independently confirmed handoff found
within the finite search budget, not a globally shortest possible intervention.
The omitted actions were already generated to verify the teacher recovery;
these reductions measure teacher supervision removed, not generation saved.
Generation costs, including discarded continuations, remain accounted for
separately in Table~\ref{tab:generation_overhead}.

\end{document}